\documentclass[10pt,twocolumn,letterpaper]{article}

\usepackage{cvpr}              

\usepackage{graphicx}
\usepackage{amsmath}
\usepackage{amssymb}
\usepackage{booktabs}

\usepackage[pagebackref,breaklinks,colorlinks]{hyperref}

\usepackage[capitalize]{cleveref}
\crefname{section}{Sec.}{Secs.}
\Crefname{section}{Section}{Sections}
\Crefname{table}{Table}{Tables}
\crefname{table}{Tab.}{Tabs.}

\def\cvprPaperID{*****} 
\def\confName{CVPR}
\def\confYear{2023}

\begin{document}

\title{Transferring Knowledge for Food Image Segmentation using\\Transformers and Convolutions}

\author{Grant Sinha\\
University of Waterloo\\
Waterloo, Ontario, Canada\\
{\tt\small gsinha@uwaterloo.ca}
\and
Krish Parmar\\
University of Waterloo\\
Waterloo, Ontario, Canada\\
{\tt\small k6parmar@uwaterloo.ca}
\and
Hilda Azimi\\
National Research Council Canada\\
Ottawa, Ontario, Canada\\
{\tt\small hilda.azimi@nrc-cnrc.gc.ca}
\and
Amy Tai\\
University of Waterloo\\
Waterloo, Ontario, Canada\\
{\tt\small amy.tai@uwaterloo.ca}
\and
Yuhao Chen\\
University of Waterloo\\
Waterloo, Ontario, Canada\\
{\tt\small yuhao.chen1@uwaterloo.ca}
\and
Alexander Wong\\
University of Waterloo\\
Waterloo, Ontario, Canada\\
{\tt\small alexander.wong@uwaterloo.ca}
\and
Pengcheng Xi\\
National Research Council Canada\\
University of Waterloo, Ontario, Canada\\
{\tt\small pengcheng.xi@nrc-cnrc.gc.ca}
}
\maketitle

\begin{abstract}
    Food image segmentation is an important task that has ubiquitous applications, such as estimating the nutritional value of a plate of food. Although machine learning models have been used for segmentation in this domain, food images pose several challenges. One challenge is that food items can overlap and mix, making them difficult to distinguish. Another challenge is the degree of inter-class similarity and intra-class variability, which is caused by the varying preparation methods and dishes a food item may be served in. Additionally, class imbalance is an inevitable issue in food datasets. To address these issues, two models are trained and compared, one based on convolutional neural networks and the other on Bidirectional Encoder representation for Image Transformers (BEiT). The models are trained and valuated using the FoodSeg103 dataset, which is identified as a robust benchmark for food image segmentation. The BEiT model outperforms the previous state-of-the-art model by achieving a mean intersection over union of 49.4 on FoodSeg103. This study provides insights into transfering knowledge using convolution and Transformer-based approaches in the food image domain.
\end{abstract}

\section{Introduction}
\label{sec:intro}

\begin{figure}[h]
    \centering
    \includegraphics[width = 0.45 \textwidth]{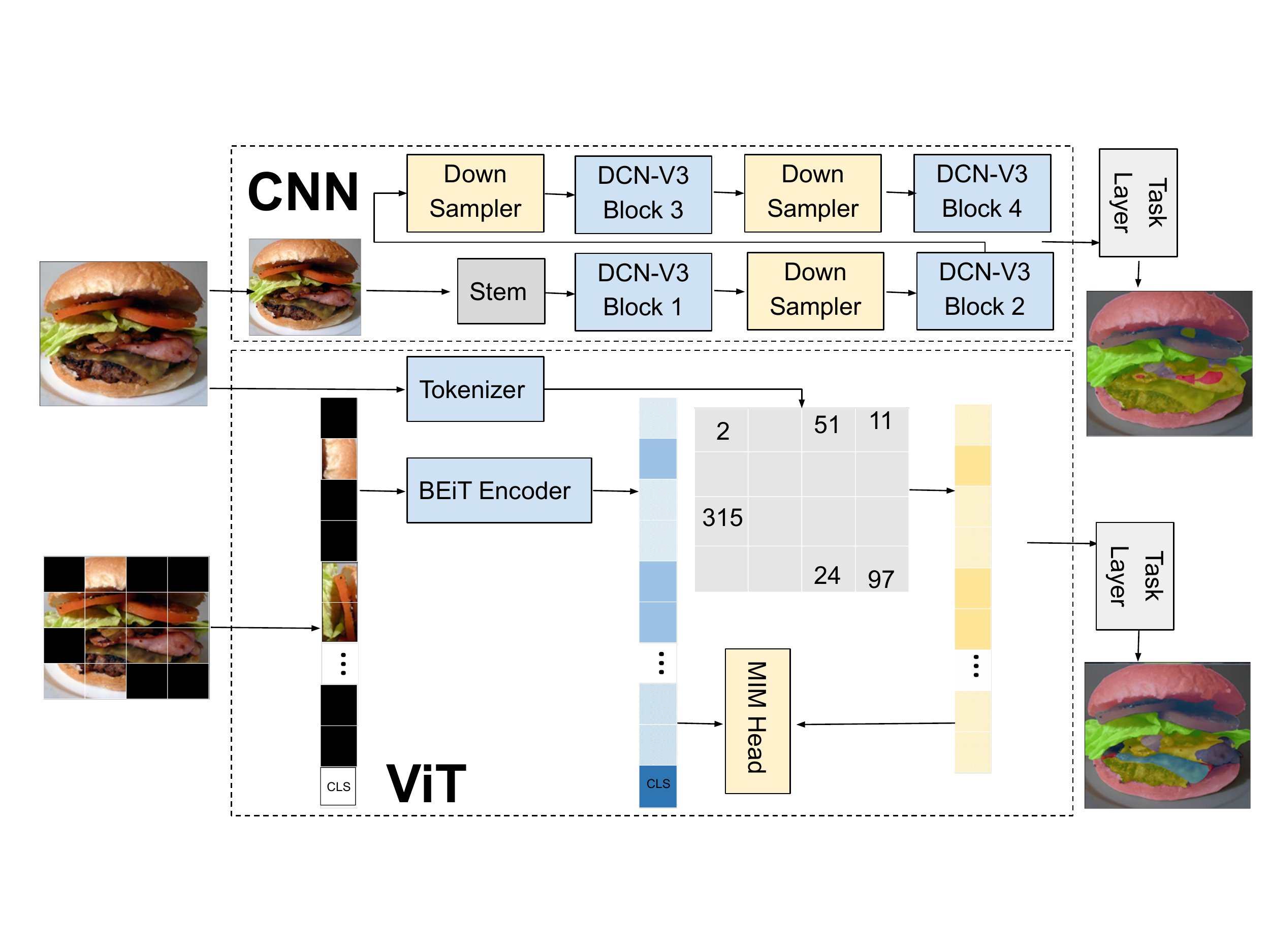}
    \caption{Comparison of downstream task performance of CNNs and ViTs on food image segmentation for the purpose of evaluating feature transferability.}
    \label{fig:header}
\end{figure}

Every year, malnutrition causes a \$10 billion annual burden on the healthcare system in the United States \cite{goates_economic_2016}. There is a strong link between poor nutrition and disease, mortality, and impaired quality of life \cite{thomas_nutritional_2000, pirlich_nutrition_2001}. Older adults are at the greatest risk of health complications due to nutritional deficiencies, being over four times more likely to be hospitalized due to malnutrition. Further, they are at higher risk because one in four adults aged 65 years or older is malnourished \cite{kaiser_frequency_2010, lanctin_prevalence_2021}.

Nutrition monitoring has been proposed as a solution for combating malnutrition among the at-risk aging population \cite{pfisterer_enhancing_2021}. Methods exist for monitoring and measuring nutritional intake, but most of them are not reliable or accurate \cite{prentice_evaluation_2011}. For example, food diaries, questionnaires, weighing, and photography each suffer from slowness, the need for trained personnel, or reporting biases. 

Recognizing and breaking down the nutritional contents of a plate of food is one of the core tasks in food imaging for nutrition monitoring. Various pipelines using computer vision models have been designed \cite{thames_nutrition5k_2021, wu_large-scale_2021}. One promising approach \cite{pfisterer_automated_2022} combines semantic segmentation with depth imaging to generate segmentation masks. The depth map is then used to compute estimates for food volumes. 


Convolutional Neural Networks (CNN) are one of the mainstream methods for semantic segmentation. CNNs are mostly lightweight in terms of computational and memory requirements and benefit from strong inductive biases. Numerous variations of CNNs have been proposed, which introduce advanced architectural designs and complex convolutional operations. For example, deformable convolutions endow CNNs with adaptive receptive fields \cite{dai_deformable_2017}, and pyramid pooling modules allow CNNs to better make use of global context clues in the scene \cite{zhao_pyramid_2017}.

More recently, Transformer-based models have achieved great success across computer vision tasks including semantic segmentation \cite{dosovitskiy_image_2021}. Equipped with multi-headed self-attention mechanisms, these models have a global receptive field that enables them to make better use of global context information, although this comes at the expense of a higher computational and memory cost \cite{vaswani_attention_2017}.

In addition to the ubiquity of Transformer-based models, recent years have also seen a shift towards large-scale vision models. BEiT \cite{bao_beit_2022, peng_beit_2022} and InternImage \cite{wang_internimage_2022} are both representatives of this paradigm. They are large-scale models pre-trained on ImageNet and capable of being fine-tuned for numerous downstream vision tasks \cite{fang_eva_2022}. After pre-training, a task-specific layer or network is appended, and the whole network is fine-tuned for the desired downstream task. 

Semantic segmentation on food images presents a unique set of challenges. One of them is that foods can occlude and mix with one-another in complex ways, preventing the model from accurately deliminating between them \cite{pfisterer_enhancing_2021}. Another challenge is the presence of inter-class similarity and intra-class variability. For example, chicken and ham can be prepared to have a very similar appearance. Different cooking techniques applied to chicken can result in vastly different visual characteristics. An effective food segmentation model must be able to recognize the diverse appearances of chicken without confusing it with other similar-looking foods like ham \cite{wu_large-scale_2021}.

Food image datasets are not as large or robust as other domains, especially when it comes to the need of image masks for training semantic segmentation models. UBIMIB2016 \cite{ciocca_food_2017}, UECFoodPixComplete \cite{del_bimbo_uec-foodpix_2021}, and FoodSeg103 \cite{wu_large-scale_2021} are datasets with fine-grained segmentation masks. Other existing datasets are mainly for food recognition and coarse-grained segmentation tasks. For instance, Recipe1M comprises one-million food images with corresponding recipes \cite{marin_recipe1m_2019}, and Food101 contains approximately one-hundred thousand food images across 101 classes \cite{lukas_food-101_2014}. However, neither of them includes annotation masks for segmentations. Among the food datasets that support semantic segmentation, we identify FoodSeg103 as the most robust \cite{wu_large-scale_2021}. This is due to its detailed, pixel-level annotations, plus empirical evidence of its relatively high difficulty. Previous works have achieved a mean intersection over union (mIoU) over all classes of up to 45.1 using a ViT-B/16-based Segmentation Transformer \cite{zheng_rethinking_2021, wu_large-scale_2021}. 


In this study, we aim to gain insight into how convolutional and Transformer-based architectures differ in knowledge transfer to the food image domain. Convolution-based methods are one of the main-stream approaches for segmentation, including the recent InternImage model \cite{wang_internimage_2022}, which is based on a deformable convolution operation, DCN-V3. It has achieved state-of-the-art results on the challenging ADE20k dataset for semantic segmentation \cite{zhou_scene_2017}. Therefore, we include InternImage \cite{wang_internimage_2022}  as a baseline model. 

In this work, we propose utilizing the BEiT v2 model \cite{peng_beit_2022}, which is a recent Transformer-based model that has yet to be applied to food image segmentation. The BEiT v2 encoder is more sophisticated than prior works, owing to its pre-training procedure, which involves training an image tokenizer and visual codebook (see Figure \ref{fig:header}). This promotes the learning of richer, semantic-level information, rather than lower-level, pixel-based information. 


Through exploring the distinctions between InternImage and BEiT v2 on knowledge transfer to the domain of food images, we make the following findings and contributions:
\begin{itemize}
    \item We survey the landscape of food image datasets for the purpose of semantic segmentation. We identify the most robust dataset and explain what makes it strong for semantic segmentation. Additionally, we explain why other datasets may not be suitable for this task.
    \item We train a tokenizer to investigate the power of vector-quantized knowledge distillation in the food image domain. We find that the tokenizer is learning semantic concepts in the food image dataset.
    \item We evaluate state-of-the-art models in the new domain of food images. Our experiments will provide insight into how knowledge transferability differs between convolution and Transformer-based architectures.
    \item We establish top results on the FoodSeg103 dataset. In particular, a BEiT v2 model achieves a mIoU of 49.4 on FoodSeg103, outperforming the previous state-of-the-art model's mIoU of 45.1.
\end{itemize}
 
\section{Methods}
\label{sec:methods}

\subsection{Dataset}

\begin{figure}
\subfloat[UNIMIB2016]{\includegraphics[width=1.055in]{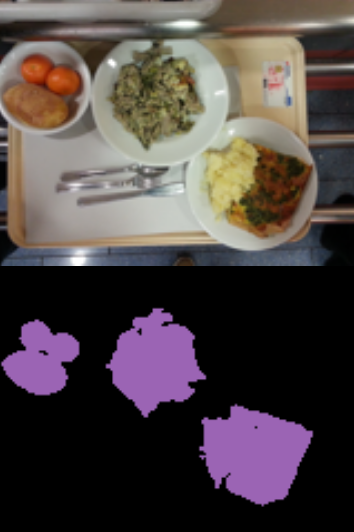}}\hfill
\subfloat[UECFoodPixComplete]{\includegraphics[width=1.065in]{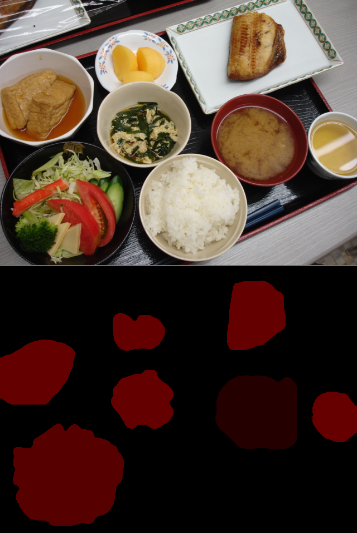}}\hfill
\subfloat[FoodSeg103]{\includegraphics[width=1.05in]{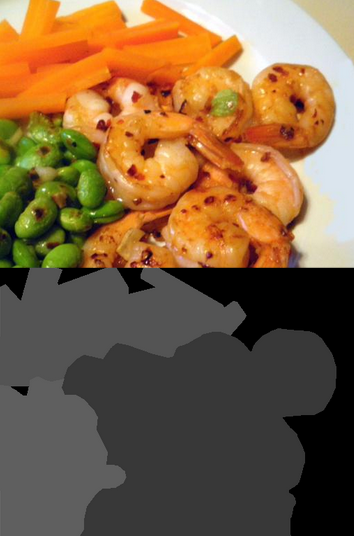}}
\caption{Image, mask pair from UNIMIB2016, UECFoodPixComplete, and FoodSeg103.}
\label{fig:imgmasks}
\end{figure}

The field of food image datasets has expanded in recent years, yet few datasets are specifically designed for semantic segmentation. Furthermore, each of these datasets has its own unique strengths and weaknesses. The datasets we investigate for semantic segmentation include UNIMIB2016, UECFoodPixComplete, and FoodSeg103 \cite{ciocca_food_2017, del_bimbo_uec-foodpix_2021, wu_large-scale_2021}. We choose these datasets because they all provide segmentation masks. Larger food image datasets exist but they do not contain segmentation masks. We include a sample image and mask pair from each of UNIMIB2016, UECFoodPixComplete, and FoodSeg103 in Figure \ref{fig:imgmasks}.

UNIMIB2016 was one of the first datasets in the food domain with ground truth food masks to support food segmentation. It was published in 2017 by researchers from the University of Milano-Bicocca for the purpose of investigating food segmentation, food consumption, and dietary monitoring. The dataset consists of 1027 food images spanning 73 food categories. Food image segmentation masks are created manually with polygonal bounding boxes. Further, masks are not per-item, so all food classes are clubbed into one category in the segmentation masks \cite{ciocca_food_2017}. 

UECFoodPixComplete was released in 2020 by researchers from the University of Electro-Communications. It includes 102 dishes and comprises 10,000 food images. The segmentation masks were obtained semi-automatically using GrabCut, which segments images based on user-initialized seeding \cite{rother_grabcut_2004}. The automatically-generated masks were further refined by human annotators based on a set of predefined rules \cite{del_bimbo_uec-foodpix_2021}.

FoodSeg103 is a recent dataset designed for food image segmentation, consisting of 7,118 images depicting 730 dishes. It was released in 2021 and the dataset's masks are pixel-level. They were obtained through manual annotations. The masks were then inspected for further refinement. Compared to UECFoodPixComplete, FoodSeg103 proves to be a more challenging benchmark for food image segmentation. In experiments conducted with a deeplabv3+ model \cite{deeplabv3plus2018}, a lower mIoU of 34.2 was achieved on FoodSeg103 than UECFoodPixComplete \cite{wu_large-scale_2021}. Further, unlike UECFoodPixComplete, which covers entire dishes but lacks fine-grained annotation for individual dish components, FoodSeg103 aims to annotate dishes at a more fine-grained level, capturing the characteristics of each dish's individual ingredients.

We conclude that FoodSeg103 is the most suitable dataset for training and evaluating a model for food image segmentation. In Table \ref{table:datasets}, we summarize various datasets, including some larger datasets that lack annotation masks, which are essential for training segmentation models using a supervised paradigm. 

\begin{table}
\centering
\caption{A summary of several robust food datasets, their number of classes, number of images, and whether it includes masks for segmentation ground truths.}
\label{table:datasets}
\begin{tabular}{l c c c} 
\toprule
Dataset & Classes & Images & Masks \\ [0.5ex] 
\midrule
UNIMIB2016 & 73 & 1027 & Yes \\
UECFoodPixComplete & 102 & 10,000 & Yes \\
FoodSeg103 & 103 & 7118 & Yes\\
Recipe1M & & 887,706 & No\\
Food101 & 101 & 101,000 & No\\
\midrule
\end{tabular}
\end{table}

\subsection{InternImage}
Neural networks that make use of convolutions have been a longstanding standard in computer vision. The new InternImage model, which uses a variation of convolutions called Deformable Convolution V3 (DCN-V3), has achieved state-of-the-art results in semantic segmentation on notable benchmarks such as ADE20k \cite{zhou_scene_2017}. The DCN-V3 operation uses a $3\times3$ convolution kernel with learnable receptive fields and modulation scalars \cite{wang_internimage_2022}. By equipping convolutions with a learnable receptive field, they can overcome their weaknesses. In particular, they can learn to utilise long-range dependencies and adaptive spatial aggregation from the data. Further, InternImage also retains the benefits of computational and memory efficiency that most convolutional models have over Transformer-based models.



InternImage processes images in several stages using blocks (see Figure \ref{fig:header}). The input is down-scaled between the blocks so that each feature map has a different resolution. Each DCN-V3 block consists of a number of basic blocks, where the basic block is designed to resemble a Vision Transformer, at a high level. The basic block consists of two sub-layers. The first uses the DCN-V3 operation as its core operator, and the second is a simple feedforward network. Both outputs have layer normalization applied. Finally, each sub-layer has a residual connection surrounding it \cite{wang_internimage_2022}. InternImage is trained on ADE20k and equipped with Mask2Former (a unified framework for all segmentation tasks based on masked attention) \cite{cheng_masked-attention_2022}. For the task layer, an UperNet decoder is appended for semantic segmentation.

\subsection{BEiT}

\begin{table}
\centering
\caption{A summary of the model versions considered in experiments, including their sizes and input resolutions.}
\label{table:models}
\begin{tabular}{l c c} 
\toprule
Model Name & Number of Parameters & Crop Size \\ [0.5ex] 
\midrule
BEIT Large\hspace{0.8cm}    & 441M  & 640$\times$640 \\
BEiT v2 Base                & 163M  & 512$\times$512 \\
BEiT v2 Large               & 441M  & 512$\times$512 \\
BEiT-3                      & 1B    & 896$\times$896 \\
InternImage-B               & 128M  & 512$\times$512 \\
InternImage-L               & 256M  & 640$\times$640 \\
InternImage-XL              & 368M  & 640$\times$640 \\
InternImage-H               & 1.31B & 896$\times$896 \\
\midrule
\end{tabular}
\end{table}

The BEiT models \cite{peng_beit_2022, bao_beit_2022, wang_image_2022} are Vision Transformer-based architectures capable of encoding images. They use masked image modeling as the pre-training task. In this process, images with corrupted patches are fed into a vision encoder, with the reconstruction target being visual tokens obtained from an image tokenizer.

There are three versions of the BEiT model. The first version introduces masked image modeling as a pre-training task for ViTs, using a pre-trained tokenizer from DALL-E \cite{peng_beit_2022, ramesh_zero-shot_2021}. The second version improves upon the first by training its own visual tokenizer using ImageNet-1K \cite{bao_beit_2022}. Finally, the third version is multi-modal and significantly scales up the number of parameters (1B parameters used on vision tasks) \cite{wang_image_2022}.

We chose to use BEiT v2 for several reasons. Although BEiT v3 achieves stronger results via scale and multi-modality, its approach to vision is not upgraded from BEiT v2. We also selected BEiT v2 over the first BEiT due to its learned tokenizer, which provides a richer reconstruction target for the encoder than the pre-trained tokenizer used in the first BEiT model. The BEiT v2 model uses a teacher model to train a tokenizer using vector-quantized knowledge distillation \cite{bao_beit_2022}. The tokenizer consists of a Transformer encoder and quantizer. The quantizer works by mapping each vector output from the Transformer encoder to its nearest neighbour in a codebook, which serves as a visual vocabulary. In this way, the continuous semantic space that the Transformer maps to is quantized into discrete codes. A decoder then learns to construct teacher model outputs from the sequence of codebook vectors. Through this, the tokenizer learns a meaningful codebook of semantic concepts from the training set. This is an improvement over the first version of BEiT, which used a pretrained tokenizer from DALL-E \cite{ramesh_zero-shot_2021}.
\begin{table}
    \centering
    \caption{An overview of the models applied to the FoodSeg103 dataset, along with their mean intersection over union. InternImage and CCNet are based on convolutions; the remainder are on Transformers}
    \label{table:performance}
    \begin{tabular}{l c c}
    \toprule
    Model Name & Number of Parameters & mIoU \\ [0.5ex]
    \midrule
    \textbf{BEiT v2 Large}\hspace{0.8cm} & \textbf{441M} & \textbf{49.4}\\
    InternImage-B & 128M & 41.1\\
    SeTR-MLA & 711M & 45.1\\
    SeTR-Naive & 723M & 43.9\\
    Swin-S & 931M & 41.6\\
    CCNet & 381M & 35.5\\
    \midrule
    \end{tabular}
\end{table}

The aim of the tokenizer is to discretize the continuous semantic space into codebook categories. Discretizing the continuous semantic space into compact codes using the codebook allows us to create a semantic-aware tokenizer. Each code in the codebook represents a semantic concept from the dataset used (e.g., unique codes for plates, eyes, foods, etc.). This approach will promote the mask-image modeling from pixel to semantic-level learning, resulting in richer representations. In the food domain, where the same ingredient can have varied appearances and contexts in different images, these stronger representations will prove to be useful.

After training the tokenizer, a network of stacked Transformer blocks is trained as a vision encoder. The encoder is trained using the task of masked image modeling. During this process, the input image is split into patches, and some of the patches are masked using a learnable embedding. This corrupted image is then fed as a sequence into the Transformer blocks, with the goal of recovering visual codebook tokens from the input image. The tokenizer trained in the previous step is used to generate the ground truth values used by the BEiT v2 encoder in this step. Once the encoder has been trained, we append an UperNet decoder as the task layer for semantic segmentation, then fine-tune the model. 




\begin{figure*}[h!]
\centering
\subfloat[Prime Rib I]{\includegraphics[width=1.3in]{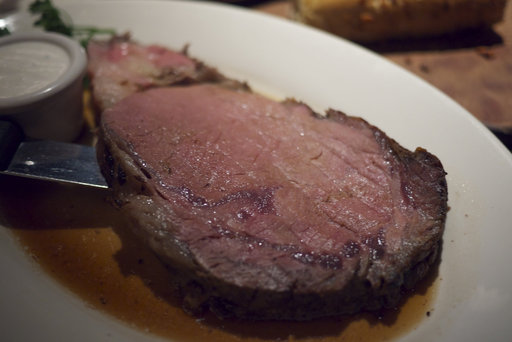}}\hfill
\subfloat[Prime Rib II]{\includegraphics[width=1.3in]{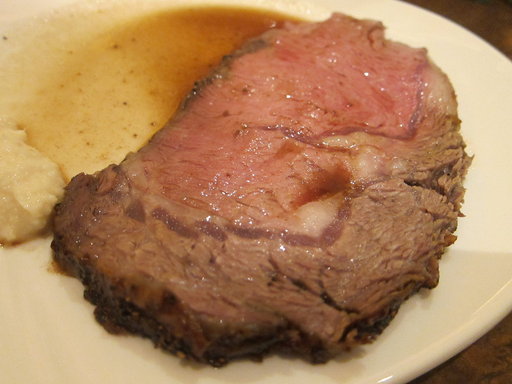}}\hfill
\subfloat[Filet Mignon I]{\includegraphics[width=1.3in]{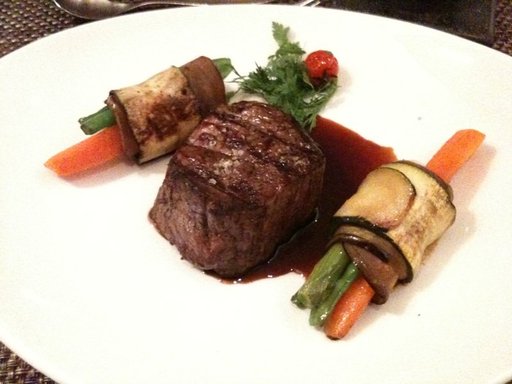}}\hfill
\subfloat[Filet Mignon II]{\includegraphics[width=1.3in]{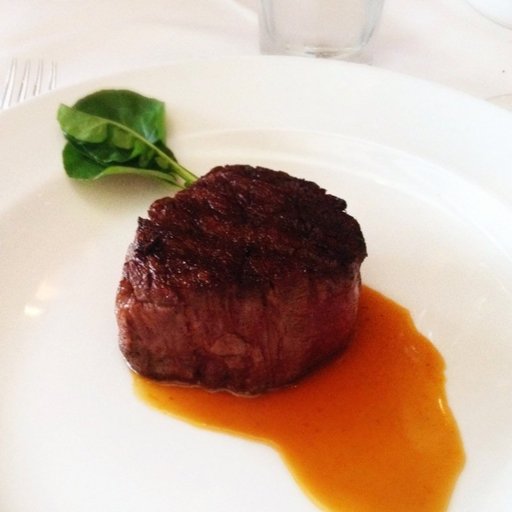}}\hfill
\subfloat[Pork Chop I]{\includegraphics[width=1.3in]{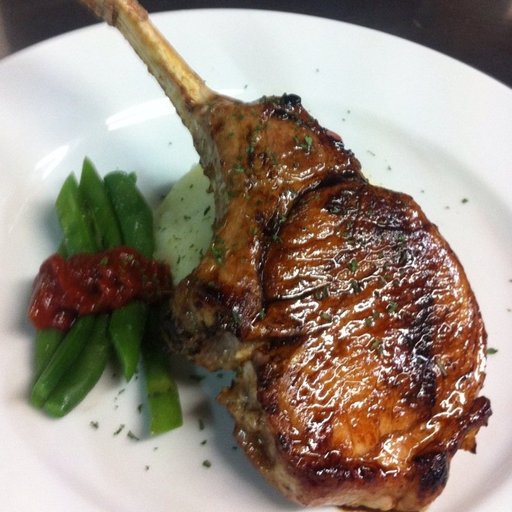}}
\caption{Food images to be tokenized using the learned tokenizer and codebook. Same-category foods have the highest IoUs, but similarly-looking cross-category foods can also yield high IoUs.}
\label{fig:tokenized}
\end{figure*}
\begin{table*}[h]
\centering
\caption{Token intersection over unions for images in Figure \ref{fig:tokenized}}
\label{table:tokenious}
\begin{tabular}{l | c c c c c} 
\toprule
& Prime Rib I& Prime Rib II & Filet Mignon I & Filet Mignon II & Pork Chop I\\
\midrule
Prime Rib I& -     & \textbf{0.281} & 0.121 & 0.153 & 0.131 \\
Prime Rib II& 0.281 & -     & 0.123 & 0.126 & 0.152 \\
Filet Mignon I& 0.121 & 0.123 & -     & \textbf{0.242} & \textbf{0.233} \\
Filet Mignon II & 0.153 & 0.126 & 0.242 & -     & 0.171 \\
Pork Chop I & 0.131 & 0.152 & 0.233 & 0.171 & -     \\ 
\midrule
\end{tabular}
\end{table*}

\section{Results}
\label{sec:results}

\subsection{InternImage Model}
Among several sizes of the InternImage model that we considered for experiments, we use the base InternImage-B model to compare against the large BEiT v2 model. They use the same crop size and have a comparable number of parameters. The relative sizes of the various InternImage and BEiT models are summarized in Table \ref{table:models}.

The InternImage encoder is pre-trained before having a task-specific layer appended for semantic segmentation. pre-training for InternImage-B is done using classification on ImageNet-1K for 300 epochs.\cite{wang_internimage_2022}.

Following the pre-training, the InternImage models are fine-tuned for semantic segmentation on FoodSeg103. In particular, we append an UperNet decoder as a task layer for semantic segmentation. Subsequently, the model undergoes end-to-end fine-tuning using the AdamW optimizer for 160K iterations. We use a learning rate of $6e-05$, no weight decay, and betas of $(0.9, 0.999)$. 

\subsection{BEiT Model}
\subsubsection{BEiT Tokenizer Training}
We trained a BEiT v2 tokenizer on Food101 to investigate the power of vector-quantized knowledge distillation. The visual tokenizer is initialized as a ViT-B/16 Transformer and taught by a CLIP-B/16 model \cite{radford_learning_2021}. The codebook size is 8192, and the dataset used is Food101, which has 101,000 images. The model is trained for 100 epochs, and the input image resolution is $224 \times 224$, which gives $14 \times 14$ total image patches each with resolution $16 \times 16$.

To evaluate the trained tokenizer, we run inference on food images from several categories in Food101, obtaining their corresponding token sequence. We compare tokenized images by considering the IoU of their sets of tokens. In particular, given tokenized images $\mathcal{I}_i$, $\mathcal{I}_j \subseteq \{1, ..., 8196\}$, we compute $$\frac{|\mathcal{I}_i \cap \mathcal{I}_j|}{|\mathcal{I}_i \cup \mathcal{I}_j|}$$ to measure the similarity of the tokenized images. We find that images belonging to the same food category typically have a higher degree of similarity, indicating that the tokenizer is successfully learning semantic concepts for the codebook. Figure \ref{fig:tokenized} and Table \ref{table:tokenious} show a collection of food images and their corresponding token IoUs. Encouragingly, we see that the highest IoUs come from foods of the same class. However, we also see the inter-class similarity of the food domain in the similarity of the tokenized pork chop and filet mignon.

\begin{figure*}
\centering
\subfloat[FoodSeg103 sample]{\includegraphics[width=1.65in]{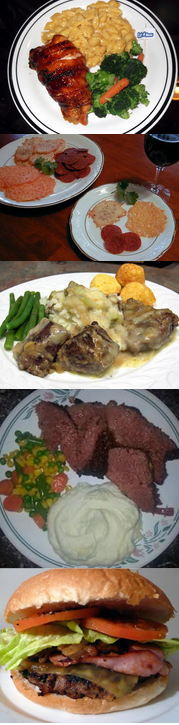}}\hfill
\subfloat[BEiT prediction]{\includegraphics[width=1.65in]{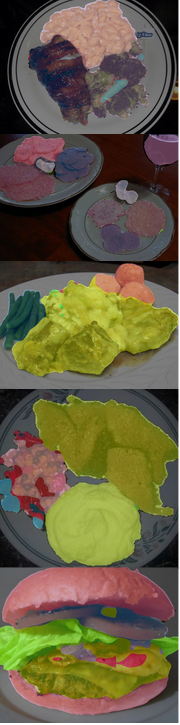}}\hfill
\subfloat[InternImage prediction]{\includegraphics[width=1.65in]{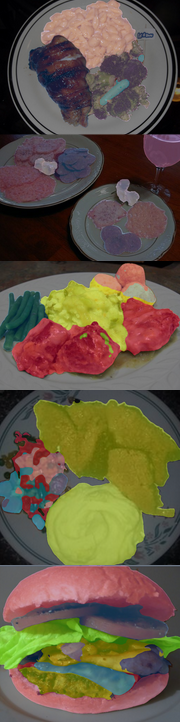}}\hfill
\subfloat[Ground Truth]{\includegraphics[width=1.65in]{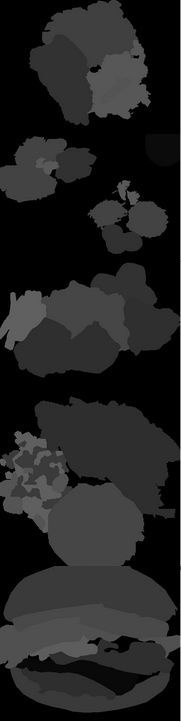}}
\caption{A comparison of inference on FoodSeg103 by the best-performing BEiT v2 and InternImage-B model.}
\label{fig:inference}
\end{figure*}

\subsubsection{BEiT model tuning}

After learning the visual tokenizer, pre-training of the BEiT v2 encoder is conducted. For the large-size model, it is initialized as a ViT-L/16 large-size Vision Transformer with a 16-pixel patch size. The tokenizer that generates ground truths for the pre-training of masked image modeling is used here. Masked image modeling is performed using ImageNet-1k at a resolution of $224 \times 224$ for 1600 epochs, resulting in $14 \times 14$ image patches. A masking ratio of 40\% is used, which means that up to 75 patches will be masked per image.


Given the pre-trained large-size BEiT v2 encoder, we append an UperNet decoder for the task layer. The entire model is then fine-tuned for 160K iterations (approximately 50 epochs) at an input resolution $512 \times 512$ using the AdamW optimizer. We use a learning rate of $3e-05$, weight decay of 0.05, and betas of $(0.9, 0.999)$. The best-performing model during training achieved a mIoU of 49.4 after 160k iterations.

\subsection{Performance of Models on FoodSeg103}
For the BEiT v2 model, during training, the best-performing model achieved a mIoU of 49.4 at iteration 160K. This is the strongest result reported on FoodSeg103, representing a new state-of-the-art for food segmentation on the dataset. On the other hand, for the InternImage-B model, during training, the best-performing model achieved a mIoU of 41.1 at iteration 48k. Table \ref{table:performance} shows the performance of these models in contrast to the other models previously applied to FoodSeg103. BEiT v2 Large outperforms the prior state-of-the-art model and uses less parameters.

Figure \ref{fig:inference} shows a visual comparison of segmentation mask predictions made by our trained models. Although both models make good segmentation prediction masks, these randomly selected testing cases prove to be challenging in terms of detecting food items. For example, on second row, the InternImage model failed to segment the food item on the top left corner. In the third example, the BEiT model failed to segment meat chunks from neighboring food items. In the last example, both models were only able to segment parts of the bacon in the burger.

\section{Discussions}
Several challenges affect the performance of the model in food image segmentation. The distribution of ingredients in food image segmentation is long-tailed, resulting in sparse data for ingredients in the long-tail, and thus, the model poorly predicts these categories. In the FoodSeg103 dataset used, we identified several problematic foods in the long-tail that the model struggled to segment correctly. hamburger appeared 7 times in the training set and 1 time in the test set, pudding appeared 5 times in the train set and 1 time in the test set, and kelp appeared 4 times in the train set and 5 times in the test set. These foods posed difficulty for each model. Figure \ref{fig:inference} shows an example of a hamburger, which both models struggle to accurately segment. 

A second challenge unique to food is the degree of inter-class similarity and intra-class variability. Each class in food images represents a food ingredient that can be prepared in numerous ways, resulting in a wide variation in appearance. This can also lead to visually similar-looking foods being mistaken for each other. This can be seen in Figure \ref{fig:tokenized} and Table \ref{table:tokenious}, where a visually similar filet mignon and pork chop are tokenized similarly. The models must be capable of understanding that a food ingredient prepared in any way is still the same food, while also differentiating between similar-looking foods \cite{wu_large-scale_2021, pfisterer_enhancing_2021, pfisterer_automated_2022}.

Owing to the superior abilities of Transformers, BEiT exhibited strong improvement over the state-of-the-art, while the performance of InternImage plateaued at a lower mIoU. One factor we can attribute this to is the global receptive field of Transformers. Even though InternImage uses learnable deformable convolutions, its kernel size is still restricted to $3 \times 3$, which prevents InternImage from having the same level of global understanding as BEiT. Global context undoubtably gives valuable insight into categorizing pixels, as certain food items are more common to occur alongside other specific food items. Another factor we may attribute BEiT v2's performance to is the use of vector-quantized knowledge distillation to train a tokenizer for BEiT's pretraining reconstruction targets. This pretraining routine endows BEiT v2 with a stronger grasp of the dataset semantics than InternImage, as the continuous, high-dimensional semantic space is quantized into codes, enabling the model to learn a visual vocabulary through its codebook.

Although BEiT outperformed InternImage in mIoU, there are instances where BEiT struggles to distinguish between classes where InternImage does not. For instance, in the third sample of Figure \ref{fig:inference}(b), BEiT groups the mashed potato and meat together, whereas InternImage was able to differentiate between them.

\section{Conclusion}
\label{sec:conclusion}

In this work, we sought to evaluate the knowledge transfer capability of Transformer and convolution-based vision backbones on the downstream task of food image segmentation. We select BEiT v2 and InternImage as strong representatives for Transformer and convolutional approaches respectively. Using these representatives and the benchmark FoodSeg103, we found that Vision Transformers have superior downstream task transferability to convolutional networks for the task of food image segmentation.


For future direction, pre-training BEiT v2 on Food101 could obtain stronger representations for food images than if it were trained on ImageNet-1K \cite{russakovsky_imagenet_2015, bao_beit_2022}. Furthermore, the multi-modal capabilities of BEiT-3 could be used to obtain even stronger food representations \cite{wang_image_2022}.

\section*{Acknowledgment}

The authors would like to acknowledge support from Aging in Place Challenge Program at the National Research Council of Canada (AiP-006).

{\small
\bibliographystyle{ieee_fullname}
\bibliography{PaperForReview}
}

\end{document}